\documentclass[runningheads]{llncs}
\usepackage{graphicx}
\usepackage{xurl}
\begin{document}
%
\title{Self-tuning hyper-parameters for unsupervised cross-lingual tokenization}
%
%
\author{Anton Kolonin\inst{1,2}\orcidID{0000-0003-4180-2870}}
\authorrunning{A. Kolonin}
%
\institute{Novosibirsk State University, Novosibirsk, Pirogova 1, 630090, Russia
\url{https://nsu.ru/} \and
Aigents,  Novosibirsk, Pravdy 6-12, 630090, Russia\\
\url{https://aigents.com/}\\
\email{akolonin@gmail.com}}
\maketitle              
\begin{abstract}
We explore the possibility of meta-learning for the language-independent unsupervised tokenization problem for English, Russian, and Chinese. We implement the meta-learning approach for automatic determination of hyper-parameters of the unsupervised tokenization model proposed in earlier works, relying on various human-independent fitness functions such as normalized anti-entropy, compression factor and cross-split \(F1\) score, as well as additive and multiplicative composite combinations of the three metrics, testing them against the conventional \(F1\) tokenization score. We find a fairly good correlation between the latter and the additive combination of the former three metrics for English and Russian. In the case of Chinese, we find a significant correlation between the \(F1\) score and the compression factor. Our results suggest the possibility of robust unsupervised tokenization of low-resource and dead languages and allow us to think about human languages in terms of the evolution of efficient symbolic communication codes with different structural optimization schemes that have evolved in different human cultures.

\end{abstract}
\section{Introduction}

Earlier studies on unsupervised tokenization learning relying on interpretable graph-based models relying on probabilistic metrics as well as metrics expressing uncertainty, such as "freedom of transition" or "transition freedom" have shown promising results, especially relying on the latter metrics applied for English, Russian and Chinese languages \cite{Kearsley,Wrenn,KoloninRamesh2022}.

However, the latest cross-lingual study carried out by previous researchers for English, Russian, and Chinese \cite{KoloninRamesh2022} has found that different languages require different hyper-parameters to be used in the process of the model building and further application of it, in order to get highly accurate tokenization output. That means, in order to obtain completely unsupervised solution for a language acquisition task, starting from the tokenization level, some extra metric is required to drive the construction of the language specific models and definition of the parameters for the application of these models without supervision from human and prior knowledge of the language itself.

To address the latter problem, in this work we explore the possibility of meta-learning for the language-independent unsupervised tokenization setup referring to the latest of prior studies \cite{KoloninRamesh2022}. We implement the meta-learning approach for automatic determination of hyper-parameters being applied to unannotated raw training text corpus, searching for human-independent fitness functions such as normalized anti-entropy, compression factor and so called "cross-split \(F1\) score", as well as additive and multiplicative combinations of the three metrics, testing them against the conventional \(F1\) tokenization score. 

\section{Experimental Approach}

\subsection{Metrics for Tuning of Hyper-Parameters}

In order to find human-independent and culture-agnostic metric for optimal segmentation of symbolic sequences involved into massive human-to-human communications, such as human languages, we attack it from from few different perspectives. First, it is "compression" view - perspective of information compressing nature of human language \cite{KIRBY201587}, which assumes that cultural evolution of linguistic structure of each of the human languages is targeting information compression goal. Second, it is "information" view - perspective of information theory where the structure of a language could be considered as driven my objective of entropy minimization across all possible messages being transmitted in this language \cite{https://doi.org/10.48550/arxiv.1905.13687}. Third, it is "concordance" view -  social and contractual perspective of the language where is is thought as a medium of enabling efficient communication between different groups minimizing discrepancy of social contexts across these social groups \cite{10.1093/acprof:osobl/9780195396171.003.0004,Usman}. 

In fact, the three perspectives above might be seen at least complementary or even equivalent. For instance, a metric measuring optimal compression can be seen from the perspective of minimized amount of entropy \cite{10.1007/3-540-55210-3_209} and respective measure. Moreover, as it is seen later in our study, we find that "compromising" capacity of a language model to provide consensus across different training corpora, which can be attributed to different social groups, is also highly correlating with the former "compressing"  and "informative" metrics. Below we define the three measures. 

\begin{figure*}[hbt!]
  \includegraphics[width=\textwidth]{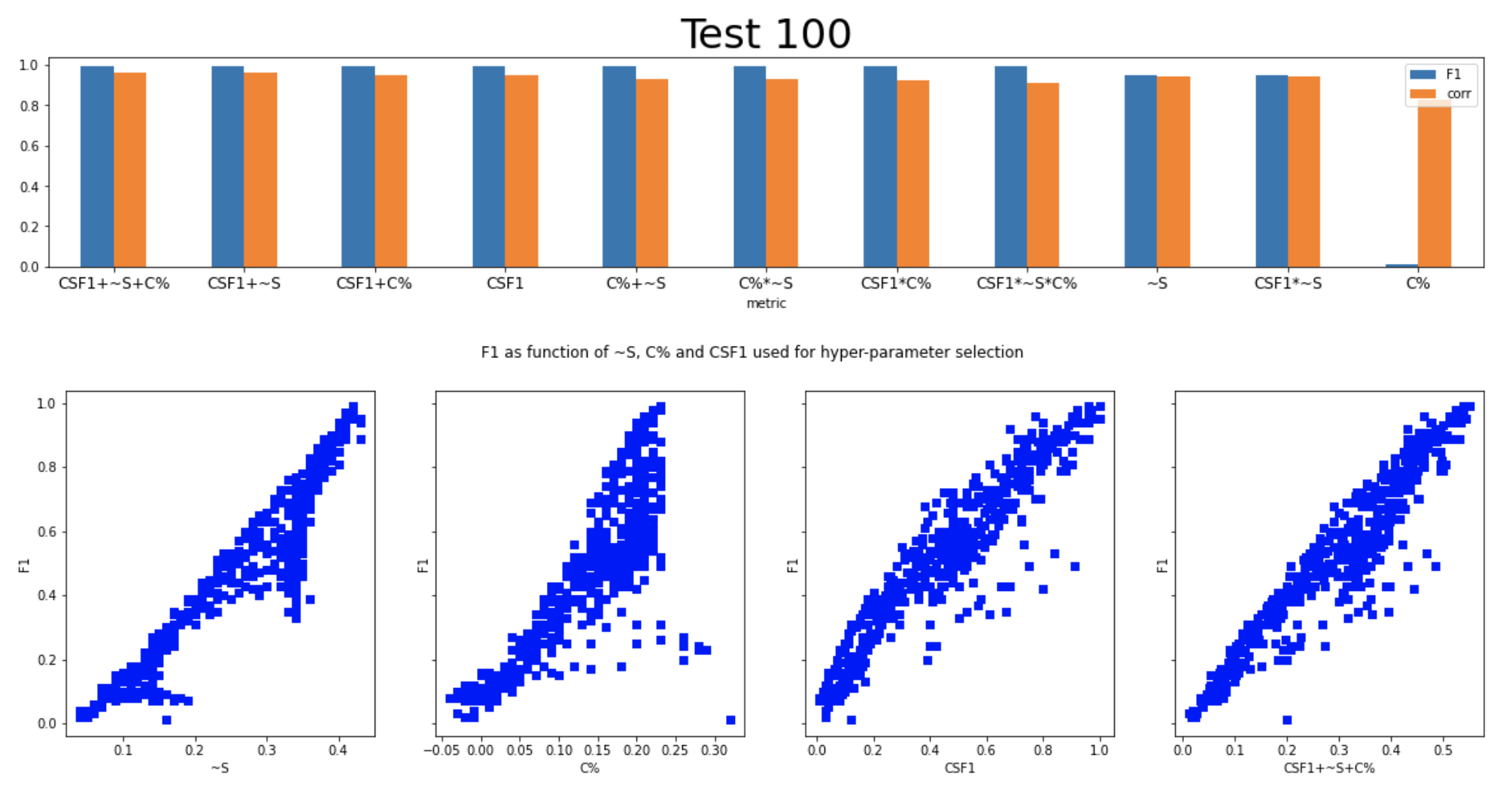}
  \caption{For the English language and test set of 100 lines. Upper row: blue bars correspond to the \(F1\) score obtained based on set of hyper-parameters corresponding to the maximum target metric on horizontal axis, blue bars correspond to Pearson correlation between the \(F1\) score and respective metric.}
  \label{fig:1}
\end{figure*}

\begin{figure*}[hbt!]
  \includegraphics[width=\textwidth]{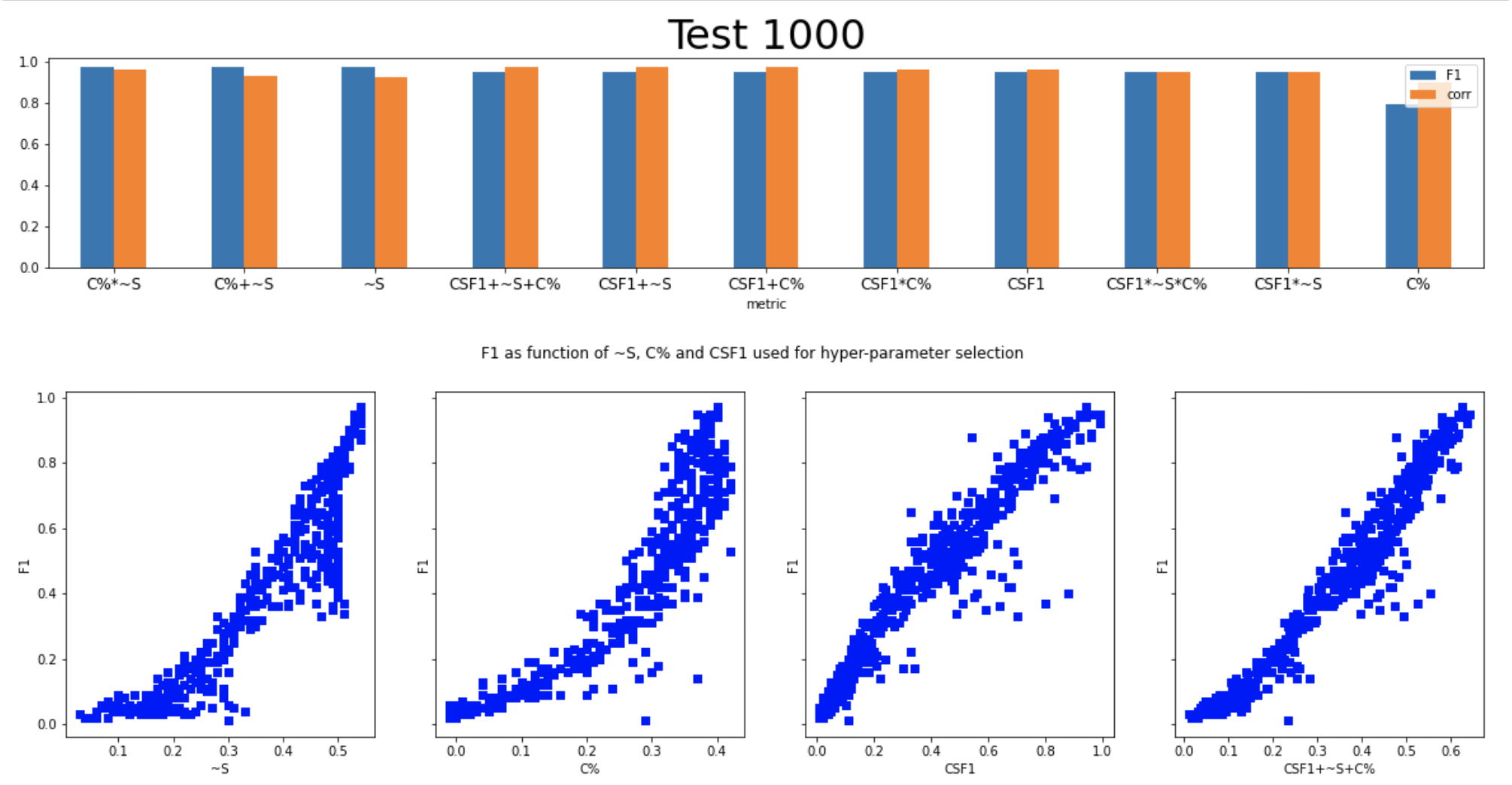}
  \caption{For the English language and test set of 1000 lines. Upper row: blue bars correspond to the \(F1\) score obtained based on set of hyper-parameters corresponding to the maximum target metric on horizontal axis, blue bars correspond to Pearson correlation between the \(F1\) score and respective metric.}
  \label{fig:2}
\end{figure*}

The "compression" tokenization metric which we call "compression factor" \(C\%\) corresponds to the ratio between numerator as "compressed" size of training set given current tokenization model and denominator as uncompressed size of training set. The compressed size is evaluated as length of sequence of token indexes in tokenized text corpus entirely plus size of the "dictionary" - sum of lengths of all tokens in it. The uncompressed size of training set is evaluated just as count of symbols in it. 

For the "information" tokenization metric we introduce so called "normalized anti-entropy" $\tilde{S}$ defined as $\tilde{S} = 1 - H/(log_2(L))$ where is the $H$ is a Shannon entropy of entire training set tokenized with given tokenization model on base $2$ and $L$ is size of the lexicon under such tokenization model.

For "concordance" metric we use what we call cross-split \(F1\) score, defined as follows. First, we split the training set corpus in two pieces of the same size, call them set \(A\) and set \(B\). Next, we create the graph traversal models across N-grams according to the previously cited work \cite{KoloninRamesh2022} for each of the corpus, call them \(M(A)\) and \(M(B)\). Then, we tokenize the test set with both models, so that \(T(M(A))\) and \(T(M(B))\) are obtained. Finally, evaluate the cross-split \(F1\) score of tokenization as  \(CSF1\) for \(T(M(A))\) against \(T(M(A))\) having one as a "ground truth" for another.

\subsection{Hyper-Parameters for Tuning}

The experimental setup involved finding conventional \(F1\) score metrics for tokenization of test set used in the referenced work \cite{KoloninRamesh2022}, using the "transition freedom" from the same work as a metric driving tokenization, given the three-dimensional grid of three hyper-parameters such as follows:
\begin{itemize}
\item The combination of $N$ ranks \(N_s\) used to perform model graph traversal and the “mean” metric computation based on a specified subset of N-grams. We have explored every possible individual value of $N$ in range from 1 to 7 (1 to 3 in case of Chinese) as well as combinations of $N$-values.     
\item Model compression threshold \(T_{mc}\) used to remove low-frequency N-grams (corresponding to vertices and transitions between them on the model graph). We have used the following values: $0.0$ (corresponding to no compression at all), $0.0001$, $0.001$, $0.01$, and $0.1$.
\item Tokenization metric threshold \(T_{tm}\): the value of a metric exceeding this level would correspond to a token boundary. We have used the following values: $0.0001$, $0.0005$, $0.001$, $0.005$, $0.01$, $0.02$, $0.05$, $0.1$, $0.2$, $0.3$, $0.4$, $0.5$, $0.6$, $0.7$, $0.8$, $0.9$.
\end{itemize}

Along with the \(F1\) scores for each point in 3-dimensional space of hyper-parameters, the target metrics \(C\%\), $\tilde{S}$ and \(CSF1\) were evaluated with the same values of hyper-parameters. We also compute end evaluate composite metrics based on the primary three ones, such as additive \(C\%+\)$\tilde{S}$ or \(C\%+\)$\tilde{S}$\(+CSF1\) and multiplicative \(C\%*\)$\tilde{S}$ or \(C\%*\)$\tilde{S}$\(*CSF1\).

\subsection{Training Sets and Model Building}

For the purpose of model building, according to the referenced work \cite{KoloninRamesh2022}, we have used public English, Russian and Chinese corpora - same as in the former paper.

For English, we have used Brown training corpus ($6$M bytes size) was downloaded from \url{http://www.sls.hawaii.edu/bley-vroman/brown$_$nolines.txt}.

For Russian, we have used RusAge corpus downloaded from \url{https://www.kaggle.com/datasets/oldaandozerskaya/fiction-corpus-for-agebased-text-classification}, having the largest one called Previews ($825$M bytes size) used for model building.

For Chinese, we have used CLUE Benchmark News $2016$ dataset as downloaded from \url{https://github.com/brightmart/nlp_chinese_corpus}. When downloaded, the folder \texttt{new2016zh} was found to have two files, \texttt{news2016zh\_valid.json} and \texttt{news2016zh\_train.json}. The latter larger file (8930M bytes size) was processed programmatically (parsing JSON; selecting \texttt{title}, \texttt{desc}, and \texttt{content} fields; and saving each of the fields as individual lines), so plain text file was produced and used for model building.

The model building has been performed following exactly the same procedure as described in the referenced work \cite{KoloninRamesh2022} with three models for each of the languages obtained.   

\subsection{Test Sets}

For the purpose of the \(F1\) score and \(CSF1\) metric determination, two test sets were involved for each of the languages.

First, there was test set of 100 text lines obtained from parallel Chinese/English corpus of 100 multi-sentence statements related to personal finance was downloaded from Magic Data (\url{https://magichub.com/datasets/chinese-english-parallel-corpus-finance}) as a tab-delimited text file with individual columns for Chinese and English versions, entitled \texttt{zh} and \texttt{en}, respectively. The corpus has been extended with Russian translations of it, with only one column entitled \texttt{ru} containing the Russian translations, as found in the file (\url{https://github.com/aigents/pygents/blob/main/data/corpora/Russian/magicdata/zh_en_ru_100/CORPUS_ZH_EN_RU.txt}) in the Aigents/Pygents open source project project.

\subsection{Experimental Approach Summary}

That is, we were able to obtain the functions \(F1(N_s,T_{mc},T_{mc})\), \(C\%(N_s,T_{mc},T_{mc})\), $\tilde{S}(N_s,T_{mc},T_{mc})$ and \(CSF1(N_s,T_{mc},T_{mc})\) and explore correlations and correspondences between them for English, Russian, and Chinese languages, for test sets of  $100$ and  $1000$ text lines relying on models and pipelines described in the cited work \cite{KoloninRamesh2022}, its appendices and code found in referenced Aigents/Pygents open source project (\url{https://github.com/aigents/pygents/}).

We were using the latter three functions of target metrics and their derivatives to find the combinations of hyper-parameters corresponding to them, and then using these hyper-parameters to perform "expectedly the best" tokenization, computing the \(F1\) score and associating the value of it with this metric.

At the same time, we were using all points in the three-dimensional space of hyper-parameters to perform tokenizations of the test sets with these parameters and compute the Pearson correlation values between the \(F1\) scores of such tokenizations and respective target and composite metrics computed based on them.  

\section{Experimental Results}

The presented table ~\ref{table:xmetrics} shows the Pearson correlations between all target metrics pairwise, for all explored combinations of hyper-parameters in the three-dimensional space of them. The table ~\ref{table:metricsF1} presents Pearson correlations between the target metrics and some of the composite metrics based on them. 

It is seen that, according to the table ~\ref{table:xmetrics}, that $\tilde{S}$ has significantly positive correlations with both \(C\%\) and \(CSF1\) across all languages. Also, all metrics are highly correlated for English and Russian, where the correlation of the $\tilde{S}$ with the other two is especially high. In turn, for Chinese, all correlations are generally lower while \(CSF1\) does not show correlation to any other metric. 

Moreover, the table ~\ref{table:metricsF1} is rendering highest correlations with \(F1\) score for composite \(C\%+\)$\tilde{S}$ while other metrics perform better for some languages and worse for others, as discussed in the language-specific sections below.   

\begin{table}
\centering
\begin{tabular}{llcc}
\hline
\textbf{Language}&\textbf{Metric1} &\textbf{Metric2} &\textbf{Correlation}\\
\hline
English & \(CSF1\) & $\tilde{S}$ & \textbf{$\mathbf{0.90}$}\\
English & \(CSF1\) & \(C\%\) & \textbf{$\mathbf{0.86}$}\\
English & $\tilde{S}$ & \(C\%\) & \textbf{$\mathbf{0.92}$}\\
\hline
Russian & \(CSF1\) & $\tilde{S}$ & \textbf{$\mathbf{0.76}$}\\
Russian & \(CSF1\) & \(C\%\) & \textbf{$\mathbf{0.64}$}\\
Russian & $\tilde{S}$ & \(C\%\) & \textbf{$\mathbf{0.83}$}\\
\hline
Chinese & \(CSF1\) & $\tilde{S}$ & \textbf{$\mathbf{0.36}$}\\
Chinese & \(CSF1\) & \(C\%\) & \textbf{$\mathbf{-0.31}$}\\
Chinese & $\tilde{S}$ & \(C\%\) & \textbf{$\mathbf{0.48}$}\\
\hline
\end{tabular}
\caption{\label{table:xmetrics}Pearson correlations between target metrics for different languages, based on larger test set of 1000 lines.}
\end{table}

\begin{table}
\centering
\begin{tabular}{llcc}
\hline
\textbf{Language}&\textbf{Metric} &\textbf{Correlation}\\
\hline
English & \(CSF1\) & \textbf{$\mathbf{0.95}$}\\
English & \(C\%\) & \textbf{$\mathbf{0.83}$}\\
English & $\tilde{S}$ & \textbf{$\mathbf{0.94}$}\\
English & \(C\%+\)$\tilde{S}$ & \textbf{$\mathbf{0.93}$}\\
English & \(C\%*\)$\tilde{S}$ & \textbf{$\mathbf{0.93}$}\\
\hline
Russian & \(CSF1\) & \textbf{$\mathbf{0.79}$}\\
Russian & \(C\%\) & \textbf{$\mathbf{0.56}$}\\
Russian & $\tilde{S}$ & \textbf{$\mathbf{0.87}$}\\
Russian & \(C\%+\)$\tilde{S}$ & \textbf{$\mathbf{0.67}$}\\
Russian & \(C\%*\)$\tilde{S}$ & \textbf{$\mathbf{0.68}$}\\
\hline
Chinese & \(CSF1\) & \textbf{$\mathbf{-0.09}$}\\
Chinese & \(C\%\) & \textbf{$\mathbf{0.85}$}\\
Chinese & $\tilde{S}$ & \textbf{$\mathbf{0.79}$}\\
Chinese & \(C\%+\)$\tilde{S}$ & \textbf{$\mathbf{0.95}$}\\
Chinese & \(C\%*\)$\tilde{S}$ & \textbf{$\mathbf{0.83}$}\\
\hline
\end{tabular}
\caption{\label{table:metricsF1}Pearson correlations between metrics and tokenization \(F1\) score for different languages, based on larger test set of 1000 lines.}
\end{table}

\subsection{English}

In case of English we find it is possible to self-tune hyper-parameters based on almost any of the metric having the \(CSF1\), $\tilde{S}$, and their composites are especially good for the purpose, referring to the table ~\ref{table:metricsF1}, figure ~\ref{fig:1} and figure ~\ref{fig:2}.

It should be noted that distributions of points on figure ~\ref{fig:1} and figure ~\ref{fig:2} corresponding to quite different test sets are pretty close and can be thought as a "signature" describing specific language model.

\subsection{Russian}

\begin{figure*}[hbt!]
  \includegraphics[width=\textwidth]{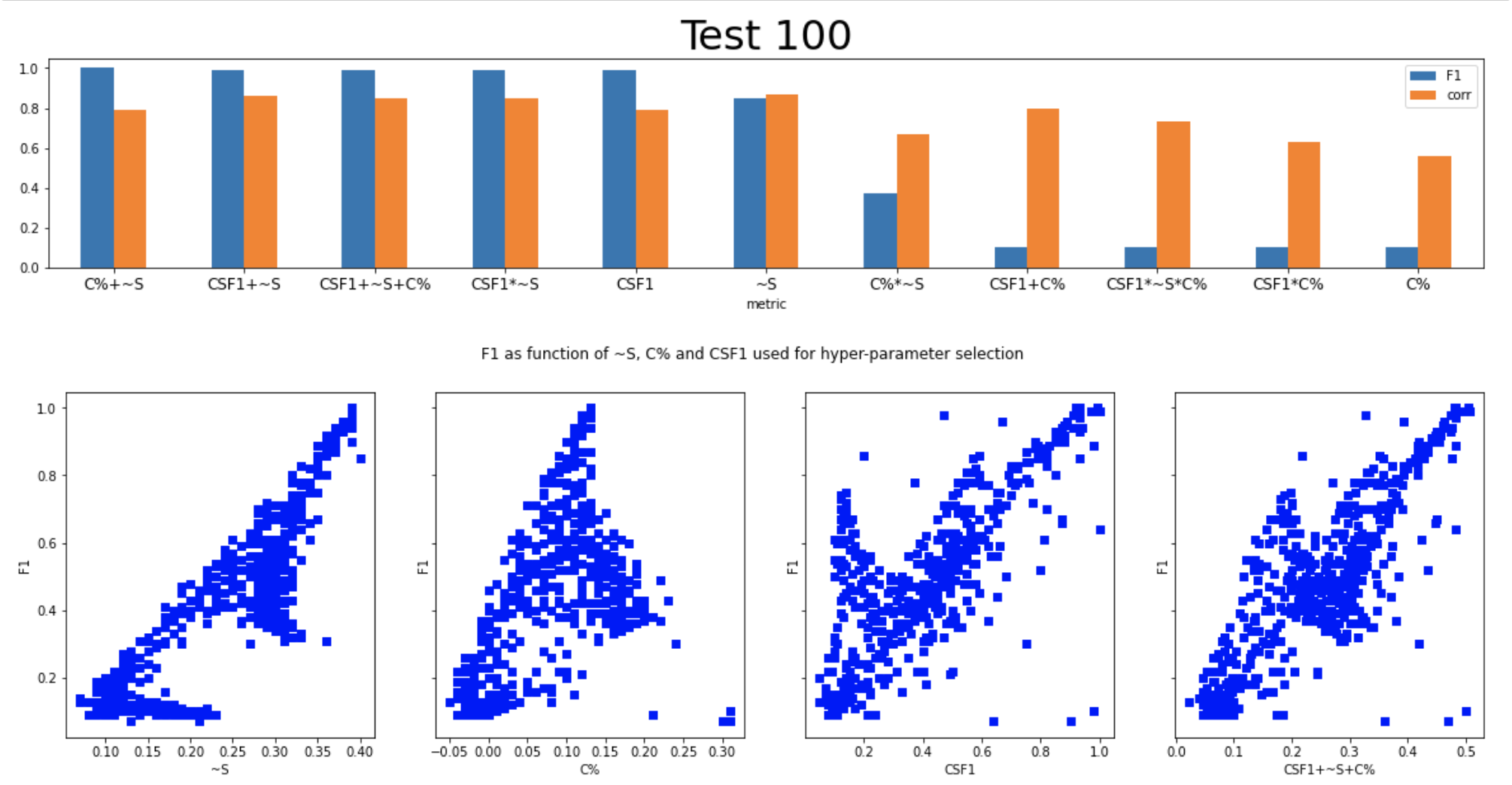}
  \caption{For the Russian language and test set of 100 lines. Upper row: blue bars correspond to the \(F1\) score obtained based on set of hyper-parameters corresponding to the maximum target metric on horizontal axis, blue bars correspond to Pearson correlation between the \(F1\) score and respective metric.}
  \label{fig:3}
\end{figure*}

\begin{figure*}[hbt!]
  \includegraphics[width=\textwidth]{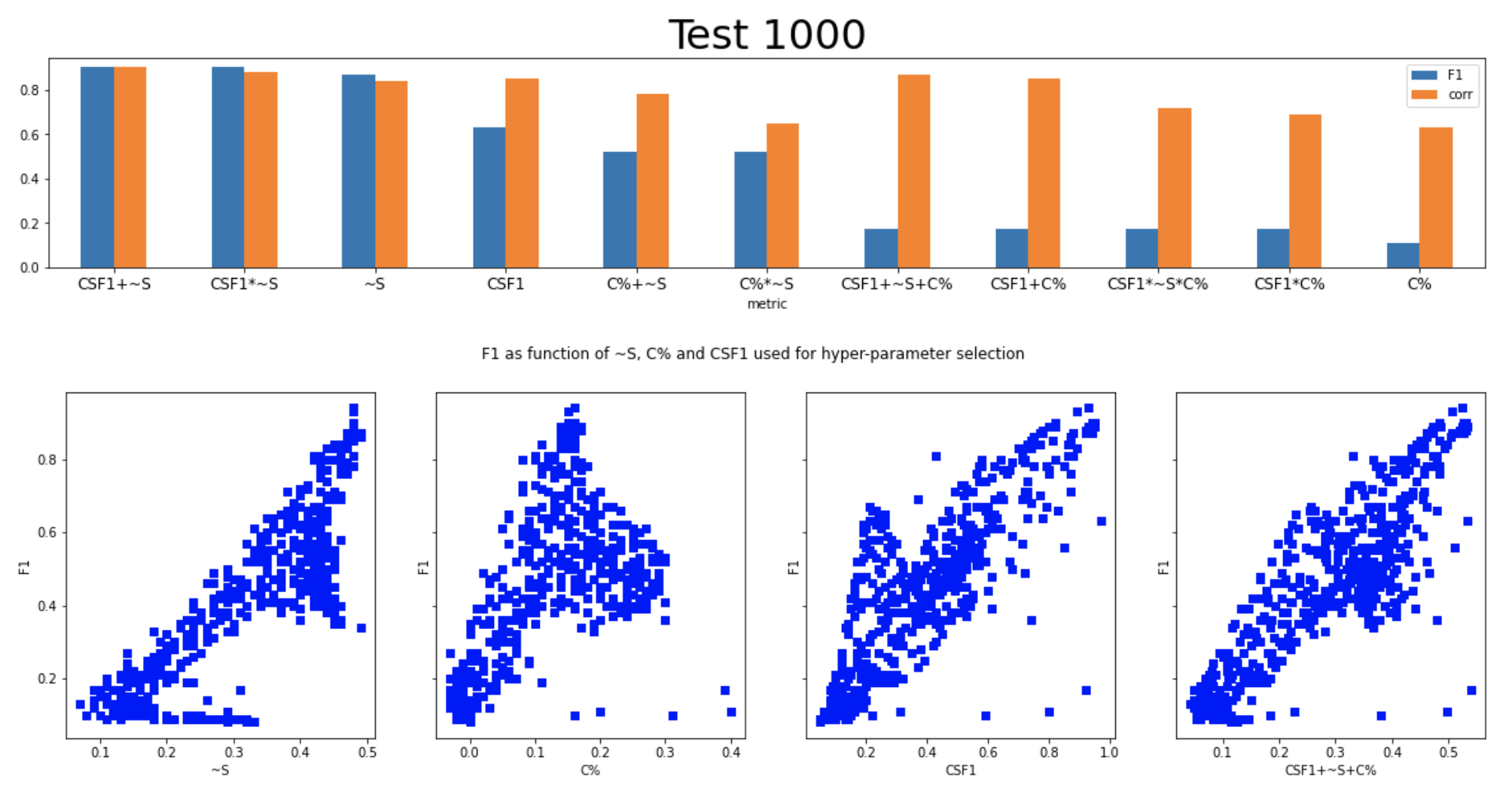}
  \caption{For the Russian language and test set of 1000 lines. Upper row: blue bars correspond to the \(F1\) score obtained based on set of hyper-parameters corresponding to the maximum target metric on horizontal axis, blue bars correspond to Pearson correlation between the \(F1\) score and respective metric.}
  \label{fig:4}
\end{figure*}

For Russian we find it substantially less reliable but still possible, relying on \(CSF1\), $\tilde{S}$, and their composites (as in case of the English), referring to the table ~\ref{table:metricsF1}, figure ~\ref{fig:3} and figure ~\ref{fig:4}.

As in case of English, distributions of points on figure ~\ref{fig:3} and figure ~\ref{fig:4} corresponding to different test sets are still quite similar pretending to serve as a "signature" of the language model.

\subsection{Chinese}

\begin{figure*}[hbt!]
  \includegraphics[width=\textwidth]{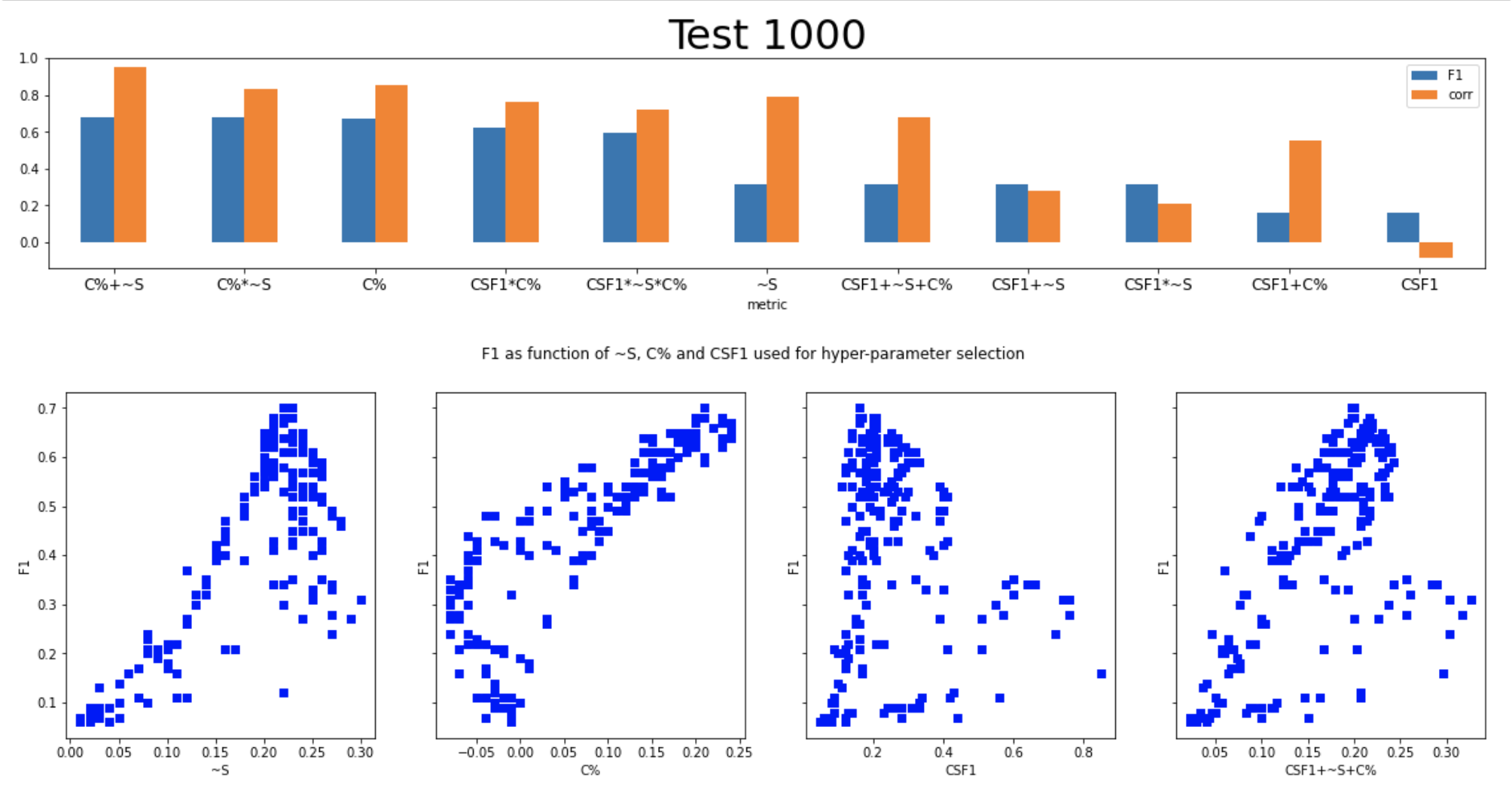}
  \caption{For the Chinese language and test set of 1000 lines. Upper row: blue bars correspond to the \(F1\) score obtained based on set of hyper-parameters corresponding to the maximum target metric on horizontal axis, blue bars correspond to Pearson correlation between the \(F1\) score and respective metric.}
  \label{fig:5}
\end{figure*}

For Chinese results are more vague than for English and Russian. However, blending the \(C\%\) with $\tilde{S}$ makes automatic determination of hyper-parameters still possible, referring to the table ~\ref{table:metricsF1}, and figure ~\ref{fig:5}. 

It should be noted that using the test set of 100 lines did not provide sufficiently reasonable result for Chinese at all and has not been presented in this paper.

\section{Conclusion}

We find a fairly good mutual Pearson correlations between all three introduced human-agnostic metrics such as "compression factor", "normalized anti-entropy" and so-called "cross-split \(F1\) score" for English and Russian languages, in the space of hyper-parameters of fully-unsupervised tokenization of raw natural language texts. We have found high Pearson correlations between these parameters and their composites against accuracy of unsupervised tokenization process performed with respective set of hyper-parameters. 

In the case of Chinese language, we find a significant correlation between the \(F1\)score and the "compression factor" and "normalized anti-entropy" as well as composites of the two. 

Our results suggest the possibility of robust unsupervised tokenization of low-resource and dead languages and allow us to think about human languages in terms of the evolution of efficient symbolic communication codes with different structural optimization schemes that have evolved in different human cultures.

At the same time it opens the way for end-to-end unsupervised language learning from unannotated and unsegmented corpora as it has been suggested by Vepstas and Goertzel \cite{https://doi.org/10.48550/arxiv.1401.3372} and further developed by Glushchenko et. al. \cite{10.1007/978-3-030-27005-6_11}.

\section*{Limitations}
The major limitation of this work appears to be that the corpora used for the three languages are way different in terms of size and topic. The English and Russian corpora are literary while the Chinese one is based on crawled news data. Further, the English corpus is 100 times smaller than the Russian one while the Chinese corpus is 10 times larger then the latter. Each of these facts or both of them might serve as an explanation of better results for English compared to Russian and Russian compared to English.

In our further work we are going to re-run the same experiment based on models built upon parallel multi-lingual corpora on the same topic and of the same size across all languages involved in the experiment.

The other direction of further work might be running the same experiment for each of the languages with training corpora of different size to see if the size of the corpora might be another hyper-parameter to drive the performance of the hyper-parameter self-tuning.

\section*{Ethics Statement}
There are no apparent negative ethical implications associated with this work. We rather suggest that our work has a positive ethical benefit as it promotes the inclusiveness of cultures based on so-called "low-resource" languages and dialects that are not easily studied using current linguistic approaches. We believe that the proposed technology can facilitate the learning of such languages by providing initial lexical vocabularies based on raw field data and thus paving the way for further learning of these languages and their grammars. Another long-term positive ethical impact comes from the "interpretable" nature of this work. Our model facilitates the transition to open, transparent and human-friendly linguistic models that can be developed for any human language and put into production, thereby preventing the potential degradation of the quality of life of "black box" NLP models. No harmful social impact from our work is anticipated.

%
%
%

\begin{thebibliography}{8}

\bibitem{Kearsley}
Logan Kearsley. A hybrid approach to cross-linguistic tokenization: Morphology with statistics. In Brigham Young University, Theses and Dissertations, 5984. 2016.

\bibitem{Wrenn}
Jesse O. Wrenn, Peter D. Stetson, and Stephen B. Johnson. An unsupervised machine learning approach to segmentation of clinician-entered free text. In Proceedings of the AMIA Annual Symposium, pages 811–5. 2007.

\bibitem{KoloninRamesh2022}
Anton Kolonin and Vignav Ramesh. Unsupervised tokenization learning. In Proceedings of the 2022 Conference on Empirical Methods in Natural Language Processing, page 3649–3664. 2022.

\bibitem{KIRBY201587}
Simon Kirby, Monica Tamariz, Hannah Cornish, and Kenny Smith. Compression and communication in the cultural evolution of linguistic structure. Cognition, 141:87–102. 2015.

\bibitem{https://doi.org/10.48550/arxiv.1905.13687}
Eugene Kharitonov, Rahma Chaabouni, Diane Bouchacourt, and Marco Baroni. Entropy minimization in emergent languages. arXiv:1905.13687 [cs.CL]. 2019.

\bibitem{10.1093/acprof:osobl/9780195396171.003.0004}
John R. Searle. Language as Biological and Social. In Making the Social World: The Structure of Human Civilization. Oxford University Press. 2010.

\bibitem{Usman}
Abdurrahman Usman, Adi Mahmud, Abdulhalim Daud, and Suratman Dahlan. Language as a social instrument. Edukasi, 18:259–276. 2022.

\bibitem{10.1007/3-540-55210-3_209}
Georges Hansel, Dominique Perrin, and Imre Simon. Compression and entropy. In STACS 92, pages 513–528, Berlin, Heidelberg. Springer Berlin Heidelberg. 1992.

\bibitem{https://doi.org/10.48550/arxiv.1401.3372}
Linas Vepstas and Ben Goertzel. Learning language from a large (unannotated) corpus. 2014. arXiv:1401.3372 [cs.CL]. 2014.

\bibitem{10.1007/978-3-030-27005-6_11}
Alex Glushchenko, Andres Suarez, Anton Kolonin, Ben Goertzel, and Oleg Baskov. Programmatic link grammar induction for unsupervised language learning. In Artificial General Intelligence, pages 111–120, Cham. Springer International Publishing. 2019.

\end{thebibliography}
%

\end{document}